\documentclass[a4paper]{article}

\usepackage{INTERSPEECH2020}
\usepackage{graphicx}
\usepackage{multirow}

\title{A language score based output selection method\\ for multilingual speech recognition}

\name{Van-Huy NGUYEN, Thi-Quynh-Khanh DINH, Truong-Thinh NGUYEN, Dang-Khoa MAC}
\address{
  Vingroup Big Data Institute, Vietnam
 }
\email{\{v.huynv22,v.khanhdtq,v.thinhnt18,v.khoamd\}@vinbdi.org}

\begin{document}

\maketitle
\begin{abstract}
The quality of a multilingual speech recognition system can be improved by adaptation methods if the input language is specified. For systems that can accept multilingual inputs, the popular approach is to apply a language identifier to the input then switch or configure decoders in the next step, or use one more subsequence model to select the output from a set of candidates. Motivated by the goal of reducing the latency for real-time applications, in this paper, a language model rescoring method is firstly applied to produce all possible candidates for target languages, then a simple score is proposed to automatically select the output without any identifier model or language specification of the input language. The main point is that this score can be simply and automatically estimated on-the-fly so that the whole decoding pipeline is more simple and compact. Experimental results showed that this method can achieve the same quality as when the input language is specified. In addition, we present to design an English and Vietnamese End-to-End model to deal with not only the problem of cross-lingual speakers but also as a solution to improve the accuracy of borrowed words of English in Vietnamese.

\end{abstract}

\section{Introduction}

One of the challenges for speech recognition systems is that the cross-lingual input in systems such as machine translation or dialog systems. Besides, for non-English speaking countries, there is another problem that is the borrowing of words in English used in daily conversations. That is why building a system that accepts multilingual input has been attracting a lot of research. Studies can be divided into two main directions. The first direction is to enhance the quality of monolingual models by leveraging or sharing resources from other languages \cite{go_2019_ml,2019_jh_ml,8114354}. The second is to build models that can accept multilingual input when spoken language changes happen at either word or utterance level. In this work, we follow the second approach to be able to use the system in real-life applications such as smart home, machine translation, or dialog systems. A system followed by the second approach usually has two main parts which are identifier and recognizer. The identifier is to detect the input language from which to select or configure the decoder model of the corresponding language for the next recognition step \cite{go_2019_ml,8114354,6289013,8268945} or do a post-evaluation \cite{Waibel2000,Niesler2006LanguageIA} to select a result from outputs of the front decoder. In this architecture, the identifier is often a statistic model and needs to be trained. Therefore, it will take more time and resources during the development and operation of the system. Motivated by this reason, in this work we propose a score to automatically select the result for the target language from outputs of a multilingual end-to-end (E2E) acoustic model using a single DNN shared for all languages. The benefit of this method is that the score is calculated simply and automatically based on language models and is performed on-the-fly during the decoding process. So the system using our method could be considered as only a single decoder, and will be more compact and less latency. We also present an approach to construct a multilingual E2E model for English and Vietnamese for not only improving the performance but also improving the accuracy of borrowed English or foreign words in Vietnamese.

The rest of this paper is organized as follows. Section 2
describes our proposed method. The experimental setup with a multilingual model for English and Vietnamese is explained in Section 3. Finally, the conclusions are presented in Section 4.
\section{Methodology}
\subsection{Selection method}
The output lattice $Lat$ of a given speech signal $x$ spoken in language $L_{i}$ can be obtained using a pretrained acoustic model $AM$. Assuming that $AM$ was trained well enough with a large dataset, so the different distance $d(s_{i},s^{*}), i=1,.., N$ should be small or in other words, $s_{i}$ should be similar to $s^{*}$, where $s^{*}$ is the ground-truth transcript, and $s_{i}$ is a sentence in top-N best paths of the lattices. $s_{i}$ could be identified by applying a language model rescoring approach. In real-life applications, $x$ is just an audio signal without any language specification, therefore the rescoring method requires an independent language model (or multilingual language model) $LM_{0}$. This can simply be constructed by combining all corpora of languages into one for training. But the perplexity of $LM_{0}$ should be high since the combined corpora would be very large \cite{1318481}. The consequence is the selected $s_{i}$ could be not the best candidate. To improve this rescoring phase, in general, we first need to identify the language $L_{i}$ that $x$ belongs to, and then rescore the lattice by a specified $LM_{i}$ which was trained using only corpus of language $L_{i}$. This approach can make a mistake if the language identifier does not work well, and the system will take more latency due to the requirement of time for the language identification model. Our proposed method will automatically select the best $s_{i}$ that does not require an identifier. It includes two steps as following:
\begin{itemize}
\item Step 1: Rescoring and selecting best-path sentences $S^{*}=\{s_{i}^{*}\}$ for each language with a given input speech $x$, where $s_{i}^{*}$ is the best-path candidate of language $L_{i}$. Each $s_{i}^{*}$ is produced as (\ref{eq_best_path}).
\end{itemize}
\begin{equation}
  s_{i}^{*}= Bestpath(Rescoring(Lat,LM_{i})); i=1,..,M
  \label{eq_best_path}
\end{equation}
where: $M$ is number of languages; $LM_{i}$ is a pretrained language model of language $L_{i}$; $Rescoring(Lat,LM_{i})$ is a rescoring function with arguments $Lat$ and $LM_{i}$.
\begin{itemize}
\item Step 2: Selecting $s_{i}^{*}$ as the output if it matches the following condition.
\end{itemize}
\begin{equation}
  \widehat{s}=argmax(P(s_{i}^{*},LM_{i})); i=1,..,M
  \label{output_selection}
\end{equation}
where: $P(s_{i}^{*},LM_{i})$ named language score is a probability value estimated by using language model $LM_{i}$ for sentence $s_{i}^{*}$.

This method is based on the following assumptions: (1) if $x$ was spoken in language $L_{i}$ then $P(s_{i}^{*},LM_{i})$ should be higher than all $P(s^{*}_{i},LM_{j}), \forall{i}\ne{j}$ due to $Count(s_{i}^{*},L_{i})>Count(s_{i}^{*},L_{j})$, where $Count(s_{i}^{*},L_{i})$ is the total number of sentences $s$ appearing in the corpus of language $L_{i}$; (2) $AM$ is highly accurate when producing the $Lat$, so that any selected $s^{*}_{i}$ from rescoring step will close to the ground-truth transcript. The first assumption is clearly true unless $x$ is very short with a few spoken words and some languages $L_{i}$ and $L_{j}$ use similar alphabets. The second one could be archived if the $AM$ was trained with a high-quality and large dataset.

\subsection{Decoding pipeline}

To apply our method, a decoding pipeline is proposed as described in Figure 1. This pipeline aims to directly produce output from input $x$ without any language identification or code-switching model to reduce the latency. Giving an acoustic model $AM$, a multilingual language model $LM_{0}$, a set of language dependent models $\{LM_{i}\}$, and an input signal $x$. The decoding includes the following steps:
\begin{enumerate}
    \item First pass decoding: $x$ is decoded by using $AM$ and $LM_{0}$ to get the lattice $Lat$.
    \item Rescoring: for each $LM_{i}$ rescore $Lat$ and select the corresponding best path to produce $S^{*}$.
    \item Output selection: the final output sentence is obtained by applying (\ref{output_selection}) to $S^{*}$.
\end{enumerate}
\begin{figure*}[t]
  \centering
  \includegraphics[width=0.7\textwidth]{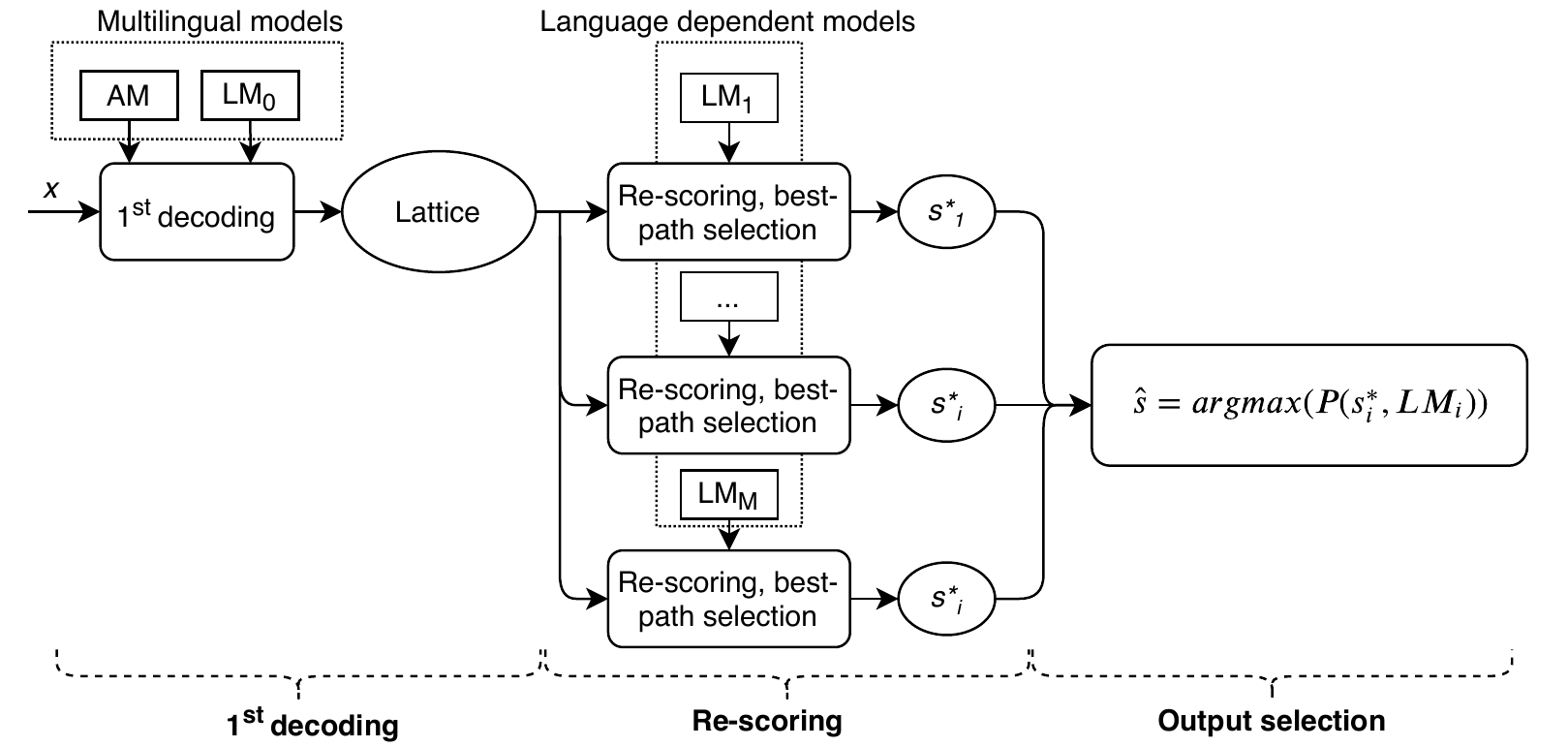}
  \caption{The multilingual decoding pipeline with the automatic output selection method}
  \label{fig:decodingpipeline}
\end{figure*}

\section{Experimental setup}
\subsection{Audio dataset}
To evaluate the proposed method, English and Vietnamese were selected for developing a multilingual model. For the acoustic model training, the training set of approximately 4000 hours audio was used. It includes 1,500 hours of English speech which were from Librispeech dataset with 360 hours \cite{librispeech}, Mozilla CommonVoice \cite{cmd_voice}, TED-LIUM corpora, and 2,500 hours from the Vietnamese corpus developed by Vingroup Big Data Institute (VinBDI-set). The Mozilla CommonVoice was used is the validated set of 780hr. This is an open corpus developed by the Mozilla Corporation. It contains hundreds of thousands of recorded voice samples with more than 40 languages around the world. The TED-LIUM corpus was the release-3 version \cite{tedlium} containing 452 hours of audio. The VinBDI-set includes two types of data. The first type is reading data. That is a recorded data from about 18,000 speakers. Each speaker was set up to randomly read 200 sentences selected from 20GB of text collected from news, online forums of popular websites in Vietnamese. All recordings were performed by using personal mobile phones and in a variety of natural environments such as restaurants, streets, offices, etc. The second type is conversations which were recorded from customer service of call centers. 

4 testing sets were used to evaluate for Vietnamese performance. They were named Reading-test, Conversation-test, YouTube-test, and VLSP2018. For English evaluation, testing sets were Test-clean and Test-other \cite{librispeech}. Those sets were randomly selected from the same raw data at the beginning to construct training, evaluation, and test sets, excepting the YouTube-test and VLSP2018. The YouTube-test was randomly collected from YouTube. It includes teenage conversations and technique talks with a lot of noises like music, non-human sounds. This was the most challenging test set since it not only is a noisy speech but also contains many foreign words. The percentage of foreign words in the vocabulary of this set is up to 23\% with 939 unique words, and most of them are common English words and property names. The accuracy of foreign words recognition will be evaluated only for this set because the ratio in others is not significant. VLSP2018 was developed by research teams involved in Vietnamese language and speech processing \cite{vlsp2018}. All audio files were saved to wave format with sample rate of 16kHz and analog/digital conversion precision of 16 bits. The summary description of these corpora is described in Table~\ref{tab:dataset}.

\begin{table*}[t]
\centering
\caption{Audio datasets}
\label{tab:dataset}
\resizebox{0.75\textwidth}{!}{%
\begin{tabular}{|l|l|l|r|r|r|c|l|}
\hline
\multicolumn{1}{|c|}{\textbf{Set}} & \multicolumn{1}{c|}{\textbf{Language}} & \multicolumn{1}{c|}{\textbf{Corpus}} & \multicolumn{1}{c|}{\textbf{Hours}} & \multicolumn{1}{c|}{\textbf{Speakers}} & \multicolumn{1}{c|}{\textbf{Sentences}} & \textbf{Foreign words} & \multicolumn{1}{c|}{\textbf{Style}} \\ \hline
\multirow{4}{*}{Training} & \multirow{3}{*}{English} & Librispeech-360hr & 363.6 & 921 & 104,014 &  & Reading \\ \cline{3-8} 
 &  & Mozilla CommonVoice & 780 & 31,858 & 644,120 & - & Reading \\ \cline{3-8} 
 &  & TED-LIUM & 452 & 2,351 & 268,263 & - & Spontaneous \\ \cline{2-8} 
 & Vietnamese & VinBDI-set & 2,500 & 18,000 & 3,666,892 & - & Reading \\ \hline
\multirow{6}{*}{Testing} & \multirow{2}{*}{English} & Test-clean & 5.4 & 87 & 2,620 & - & Reading \\ \cline{3-8} 
 &  & Test-other & 5.3 & 90 & 2,939 & - & Reading \\ \cline{2-8} 
 & \multirow{4}{*}{Vietnamese} & Reading-test & 9.9 & 23 & 5,358 & - & Reading \\ \cline{3-8} 
 &  & Conversation-test & 10.8 & 1,892 & 11,533 & - & Spontaneous \\ \cline{3-8} 
 &  & YouTube-test & 9.9 & unknown & 5,432 & 24,8\% & Spontaneous \\ \cline{3-8} 
 &  & VLSP2018 & 2.1 & unknown & 796 & - & Reading \\ \hline
\end{tabular}%
}
\end{table*}
\subsection{Universal phone set and lexicon}
In this work, the acoustic model is a multilingual model that can accept a cross-lingual input $X=\{x_{1},x_{2},..,x_{L}\}$ and convert it into a sequence of phonemes $P=``p_{1},..,p_{i},..,p_{L}"$ before passing it to the next step, where $~p_{i}$ is the phone $i^{th}$ according to speech frame $x_{i}$. The decoder will further select output words whose pronunciation transcripts that matches the $P$ based on a given pronunciation lexicon. To construct this lexicon, we created a vocabulary set by selecting all words that appeared in the transcription of the training audio data. Afterward, for English words we used the CMU dictionary \cite{cumdic} to get corresponding pronunciations, and Vietnamese words we applied a Grapheme to Phoneme (G2P) toolkit \cite{vphon}. Since this tool uses the X-SAMPA alphabet as described in \cite{2010kirby} we had to map ARPAbet phones used in CMU dictionary to X-SAMPA phones. It was based on a mapping table which is from Switchboard project \cite{switch} with the purpose that phones can be shared between languages. About 1000 of the most frequent foreign words in the collected Vietnamese corpora were selected and manually transcribed in Vietnamese reading style to improve the lexicon. To deal with the tones and stress problems in Vietnamese and English, a toneme set was created by the combination of Nucleus with tone symbols as described in \cite{7357876} and tested for E2E model in \cite{8713758}. All stress symbols in CMU dictionary phones were replaced by ``1" according to symbol ``1" in the toneme set to represent a vowel without tone. The final lexicon was a combination of all English and Vietnamese words with all their possible pronunciations. The vocabulary size was 72,000 with 6300 Vietnamese words, and the number of cross-sharing phones was 131. This vocabulary was then also used to build language models presented in the next section.

\subsection{Language models}
For English, two available prebuilt language models were used to create $LM1$. One was the unpruned Librispeech 3-gram model \cite{librispeech} trained on the source material from the Project Gutenberg books \cite{gutern}. Another was the big 4-gram model \cite{lm_kladi} trained on TED-LIUM corpus. $LM1$ was a 3-gram model produced by applying a linear interpolation method as $LM1=\alpha*LM_{Librispeech} + (1-\alpha)*LM_{TED-LIUM}$ with $\alpha$ was 0.7 (this weight yielded the best perplexities on the English testing sets in this work). The perplexities of $LM1$ on the Test-clean and Test-other were 184 and 172, respectively.

Since no official n-gram version has been published, the language model for Vietnamese was built from scratch. We used two corpora, the first of which was about 19GB \cite{vqbinh}. This corpus was collected from about 15 million articles on the internet with many different topics such as news, culture, sports, politics, etc. The second corpus, about 1GB in size, was collected from several online forums related to daily conversations about health, home electronics, and fashion to increase the number of samples related to conversation topics. The combined set of these corpora was normalized through some simple filters such as removing special characters, converting numbers to written forms, removing duplication, etc. Finally, the normalized data were combined with the transcription of the training audio data to train a 3-gram model.

After building $LM1$ and $LM2$ as above, the multilingual model $LM0$ was created by linearly interpolating $LM1$ and $LM2$ with $\alpha = 0.5$. However, $LM0$ is only used in the first decoding step to produce lattices as described in Figure~\ref{fig:decodingpipeline}, so it is not necessarily an optimal model. $LM0$ was further cut off all n-grams that have a probability of less than 2E-8 to reduce its size and speed up the first decoding step. After this step, the size of the unpruned $LM0$ from 800MB has decreased to 30MB. The perplexities of $LM0$, $LM1$, and $LM2$ on the test sets are shown in Table~\ref{tab:lms}. All of these models were built by using the tool SRILM \cite{tedlium}.

\begin{table}[t]
\centering
\caption{Language model evaluation}
\label{tab:lms}
\resizebox{0.4\textwidth}{!}{%
\begin{tabular}{|l|c|c|c|}
\hline
 & \textbf{\begin{tabular}[c]{@{}c@{}}Multilinguage\\ LM0\end{tabular}} & \textbf{\begin{tabular}[c]{@{}c@{}}English\\ LM1\end{tabular}} & \textbf{\begin{tabular}[c]{@{}c@{}}Vietnamese\\ LM2\end{tabular}} \\ \hline
\textbf{Test-clean} & 569.3 & 187.1 & - \\ \hline
\textbf{Test-other} & 522.2 & 174.3 & - \\ \hline
\textbf{Reading-test} & 136.6 & - & 87.2 \\ \hline
\textbf{Conversation-test} & 95.2 & - & 62.7 \\ \hline
\textbf{YouTube-test} & 199.5 & - & 111.4 \\ \hline
\textbf{VLSP2018} & 75.7 & - & 47.5 \\ \hline
\end{tabular}%
}
\end{table}

\subsection{Multilingual acoustic model}
\begin{figure}
    \centering
    \includegraphics[width=0.4\textwidth]{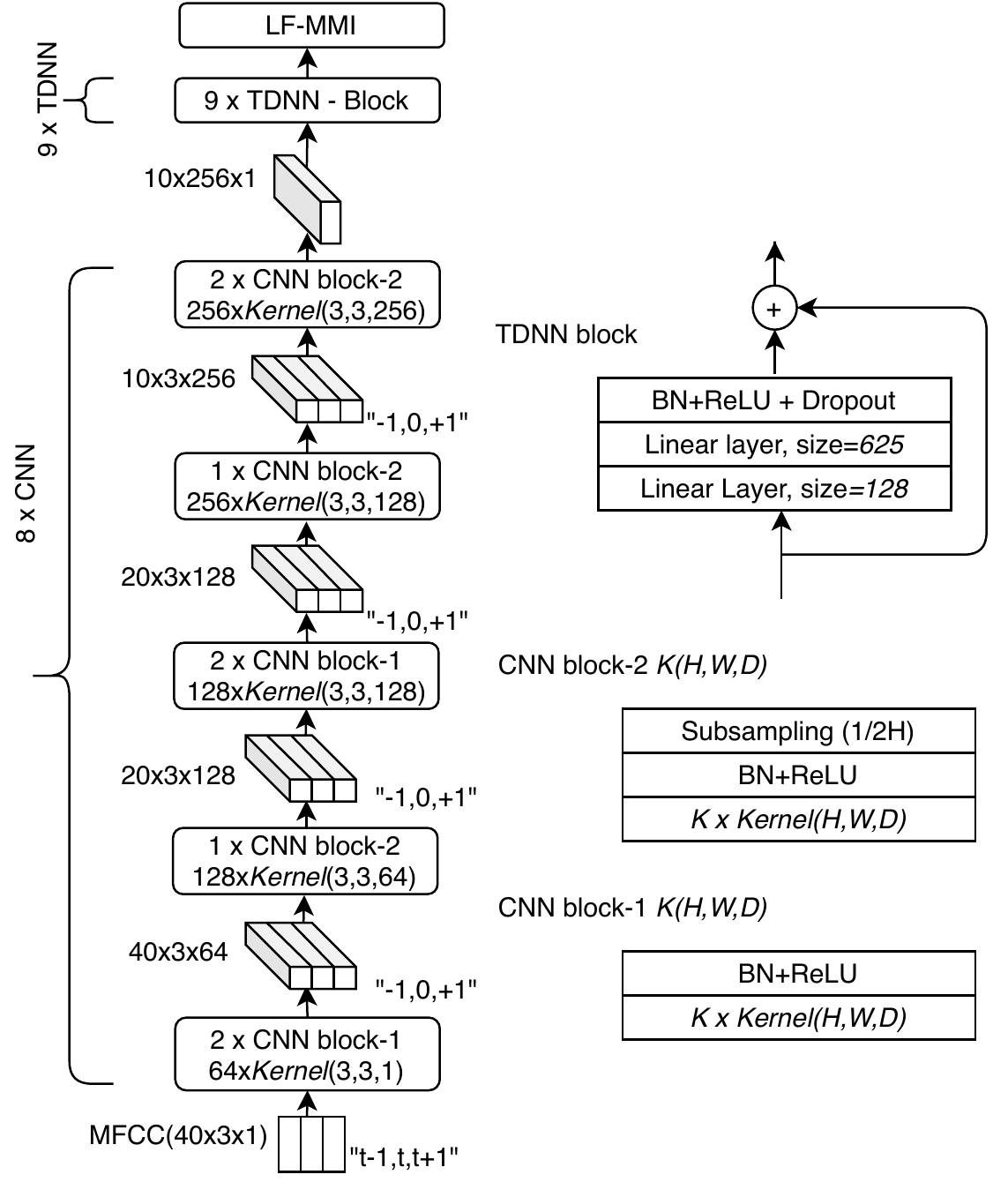}
    \caption{The multilingual acoustic Architecture}
    \label{fig:model}
\end{figure}

In this study, we propose to use a DNN architecture using an E2E training technique with the objective function of LF-MMI \cite{lfmmibasic,lfmmi}. This architecture can be divided into two parts. The first part was a stack of 8 Convolution Neural Network (CNN) layers that works as a feature extractor. Each CNN layer consists of 3 consecutive components: Convolution \cite{cnn}, Rectified Linear Unit (ReLU) and Batch Normalization (BN). The second part as a decoder was a stack including 9 layers of Time Delay Neural Network (TDNN) \cite{tdnn}. We experimented with some other architectures to chose the one with not only good WER result but also fast enough decoding that is suitable for real-time application. That is why Long Short Term Memory (LSTM) or Attention techniques were chosen within this study. The input feature before being passed into CNN layers was Mel Frequency Cepstral Coefficients (MFCCs) calculated on each speech signal frame with a size of 25ms, shifting of 10ms, and the number of coefficients was 40. The model was trained for 5 epochs using the Kaldi toolkit \cite{kaldi}. The configuration of each layer is described in Figure~\ref{fig:model}.

\subsection{Result and Discussion}
\begin{figure}[t]
    \centering  
    \includegraphics[width=0.4\textwidth]{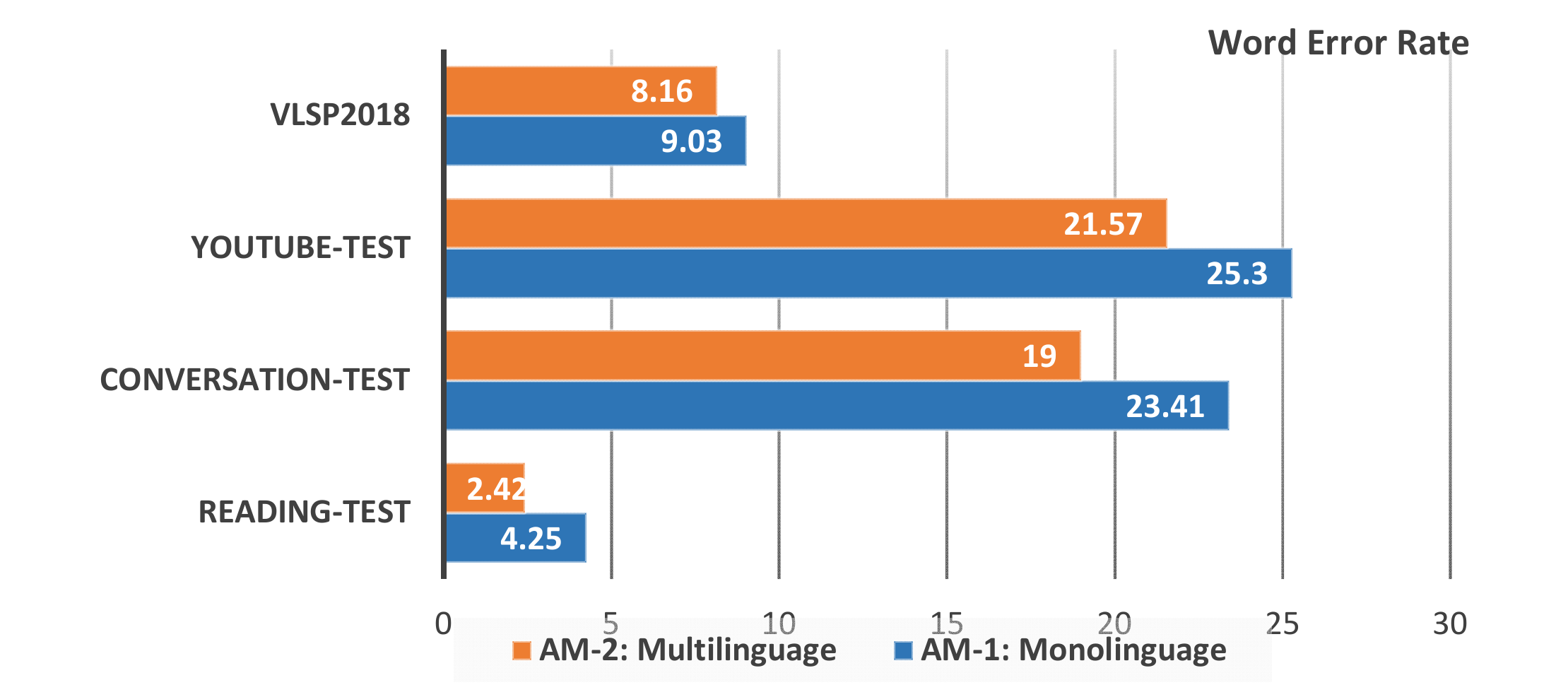}
    \caption{Comparison between monolingual and multilingual models}
    \label{fig:mono_multi}
\end{figure}

For experiments, we build two acoustic models. The first model, called AM-1, was trained using only Vietnamese datasets. This model was used as the baseline model for comparison to the second model AM-2. The AM-2 was trained on all datasets including English and Vietnamese. The models and the performance of the proposed method were evaluated by using the Word Error Rate (WER) metric measured on 6 test sets as mentioned above. Specific results are presented below.
\begin{itemize}
    \item \textbf{Monolingual vs multilingual}
\end{itemize}
For comparing AM-1 and AM-2, we measured WER on 4 Vietnamese test sets with rescored results using the Vietnamese language model LM-2. The result is shown in Figure~\ref{fig:mono_multi}. The AM-2 model, the multilingual model, on the average gave a better WER by 2.7\% on all of these test sets. This result showed that using multilingual data could improve the quality of acoustic models not only in the under-resourced cases like the vast majority of studies published on multilingual models but also for cases where large datasets are available but some acoustic domains are missed. Specifically in this work, Vietnamese training data was mostly reading and clean data. So the WERs on noisy and spontaneous speech sets like YouTube-test and Conversation-test were pretty high with the AM-1 model which was trained with this data. When training AM-2 by adding English data, we aim to supplement speech data of missing types such as spontaneous and noisy speech included in the TED-LIUM and Voicecommon sets. As a result, averagely it helped to improved WER by 4.07\% on the YouTube-test and Conversation-test sets. The correct rate of foreign words increased from 25.7\% to 58.1\% on the the YouTube-test. It demonstrated that the multilingual approach is a solution to solve the problem of foreign words. Instead of having to choose and transcribe them manually, we can take and let the models learn their pronunciations from native speakers through corresponding data. Specifically, in this work, we let the model lean foreign pronunciations from English data.
\begin{itemize}
    \item \textbf{One pass decoding with the proposed method}
\end{itemize}
To experiment that our method can automatically select the best result without having to identify the input language, we did two experiments. The first experiment was to simulate the case of the input language is known in advance. We firstly decoded all 6 testing sets to obtain lattices using AM-2 and LM0. Afterward rescored English sets by LM1 and Vietnamese sets by LM2. The final results are presented in the Table~\ref{tab:finalresult} in columns 3\textsuperscript{th} and 4\textsuperscript{th}. In the second experiment, we combined all six testing sets into one set to simulate that the decoder will not know in advance the input language. It could be English or Vietnamese at random. Lattices of this combined set were also obtained by AM-2 and LM0 at the first decoding step. But at the rescoring step, both LM1 and LM2 were used to produce the best paths and applied the proposed method to select final outputs. Finally, we separated the final outputs based on their utterance ID back to the six corresponding testing sets to individually measure WER for each one. Results are presented in column 5\textsuperscript{th} of Table~\ref{tab:finalresult}.
Comparing the results of these two experiments shown that our method produced a similar performance without a language identifier for input signals. There was a bit worse on Conversation-test and YouTube-test sets. We found that these sets contain some very short utterances with only one or two words. For these cases the language scores given by LM1 and LM2 could be similar since the output sentences chosen probably have meaning and are made up of homonyms between the two languages.
\begin{table}[t]
\centering
\caption{WER results}
\label{tab:finalresult}
\resizebox{0.4\textwidth}{!}{%
\begin{tabular}{|l|c|c|c|c|}
\hline
 & \shortstack{1\textsuperscript{st} decoding \\ with LM0} & \shortstack{Rescoring \\ with LM2} & \shortstack{Rescoring \\ with LM1} & \shortstack{Proposed \\ method} \\ \hline
Test-clean & 18.8 & - & 11.50 & 11.50 \\ \hline
Test-other & 33.35 & - & 22.59 & 22.59 \\ \hline
Reading-test & 3.84 & 2.42 & - & 2.42 \\ \hline
Conversation-test & 20.54 & 19.00 & - & 19.2 \\ \hline
YouTube-test & 24.55 & 21.57 & - & 21.83 \\ \hline
VLSP2018 & 10.12 & 8.16 & - & 8.16 \\ \hline
\end{tabular}%
}
\end{table}
\section{Conclusion}

In this work, we contributed two main points. The first is that we proposed a method to automatically select the output based on a language score when decoding with a multilingual model. By using this method the system does not need a language identification module, so the latency of the system will be improved but the accuracy is still similar to the case of input language is specified. The system will be more compatible with real-life applications where users can randomly use multiple languages during their pronunciation. Second, we presented an approach to develop a multilingual model for English and Vietnamese using a universal phone set and a single DNN architecture. The experimental results showed that a multilingual approach is a simple but effective solution to enhance a speech recognition model not only for under-resourced domains but also for foreign words. Specifically in this work, although the baseline model for Vietnamese was trained with 2500hr of data, it was still possible to reduce WER by 2.7\% when using a multilingual approach, and the improvement on foreign words was by 32.4\%. In subsequent studies we will improve the method and evaluate in more detail for cases of cross-lingual speaking happens at the word level.
\bibliographystyle{IEEEtran}
\bibliography{mybib}
\end{document}